%% file: ms.tex
\pdfoutput=1
\documentclass[11pt,letterpaper]{article} 
\usepackage{emnlp2017}
\usepackage{times}
\usepackage{latexsym}
\usepackage{amsmath}
\usepackage{amssymb}
\usepackage{booktabs}
\usepackage[]{graphicx}
\graphicspath{{figures/}}
\usepackage{tikz}
\usepackage{tikz-dependency}
\usetikzlibrary{arrows}
\usepackage{xspace}

\emnlpfinalcopy

\def\emnlppaperid{971}

\DeclareMathOperator{\dir}{dir}
\DeclareMathOperator{\lab}{lab}

\renewcommand{\vec}[1]{\mathbf{#1}}
\newcommand{\bleuone}{BLEU${_1}$\xspace}
\newcommand{\bleu}{BLEU${_4}$\xspace}

\title{Graph Convolutional Encoders\\ for Syntax-aware Neural Machine Translation}

\newcommand*{\affaddr}[1]{#1} 
\newcommand*{\affmark}[1][*]{\textsuperscript{#1}}

\author{Jasmijn Bastings\affmark[1] \quad Ivan Titov\affmark[1,2] \quad Wilker Aziz\affmark[1]\quad\\ \textbf{Diego Marcheggiani\affmark[1]}\quad \textbf{Khalil Sima'an\affmark[1]} \\
       \affaddr{\affmark[1]ILLC, University of Amsterdam} \quad
\affaddr{\affmark[2]ILCC, University of Edinburgh}\\
        {\tt \{bastings,titov,w.aziz,marcheggiani,k.simaan\}@uva.nl}
}

\date{}
\begin{document}

\maketitle
\input{abstract.tex}
\input{introduction.tex}
\input{background.tex}
\input{model.tex}

\input{experiments.tex}
\input{relatedwork.tex}

\input{conclusion.tex}
\input{ack.tex}

\bibliography{papers2017}
\bibliographystyle{emnlp_natbib}

\end{document}

%% file: abstract.tex
\begin{abstract}
We present a simple and effective approach to incorporating syntactic structure into neural attention-based encoder-decoder models for machine translation. We rely on graph-convolutional networks (GCNs), a recent class of neural networks developed for modeling graph-structured data. Our GCNs use predicted syntactic dependency trees of source sentences to produce representations of words (i.e. hidden states of the encoder) that are sensitive to their syntactic neighborhoods. GCNs take word representations as input and produce word representations as output, so they can easily be incorporated as layers into standard encoders (e.g., on top of bidirectional RNNs or convolutional neural networks). We evaluate their effectiveness with English-German and English-Czech translation experiments for different types of encoders and observe substantial improvements over their syntax-agnostic versions in all the considered setups.
\end{abstract}

%% file: introduction.tex
\section{Introduction}
\label{sec:introduction}

Neural machine translation (NMT) is one of success stories of deep learning in natural language processing, with recent NMT systems outperforming traditional phrase-based approaches on many language pairs \cite{sennrich-haddow-birch:2016:WMT}. State-of-the-art NMT systems rely on sequential encoder-decoders~\cite{sutskever2014seq2seqnips,bahdanau15iclr} and  lack any explicit modeling of syntax or any hierarchical structure of language. One potential reason for why we have not seen much benefit from using syntactic information in NMT is the lack of simple and effective methods for incorporating structured information in neural encoders, including RNNs.
Despite some successes, techniques explored so far either incorporate syntactic information in NMT models in a relatively indirect way (e.g., multi-task learning
\cite{luong16iclr,nadejdeetal2017syntaxaware,eriguchi2017nmtrg,hashimoto2017latentgraph}) or may be too restrictive in modeling the interface between 
syntax and the translation task (e.g., learning
representations of linguistic phrases~\cite{eriguchi2016treetoseq}). 
Our goal is to provide the encoder with access to rich syntactic information but
let it decide which aspects of syntax are beneficial for MT, without placing rigid constraints on the interaction between syntax and the translation task. This goal is in line with claims that rigid syntactic constraints typically hurt MT~\cite{Zollmann:2006:SAMT,smith-eisner:2006:WMT,chiang:2010:soft}, and, though these claims have been made in the context of traditional MT systems, we believe they are no less valid for NMT.

\begin{figure}[tb]
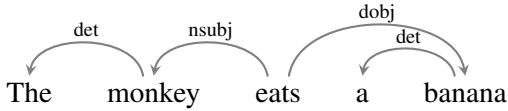

\begin{center}

    \begin{dependency}[arc edge, arc angle=80, text only label, label style={above},edge style={gray,thick}]
    \begin{deptext}[column sep=0.55cm]
    The \& monkey \& eats \& a \& banana \\
    \end{deptext} 
    \depedge{2}{1}{det}
    \depedge{3}{2}{nsubj}
    \depedge{3}{5}{dobj}
    \depedge{5}{4}{det}
    \end{dependency}
\end{center}
\caption{A dependency tree for the example sentence: ``{\it The monkey eats a banana}.'' }
\label{fig:deptree}
\end{figure}

Attention-based NMT systems~\cite{bahdanau15iclr,luong15emnlp} represent source sentence words as latent-feature vectors in the encoder and use these vectors when generating a translation. Our goal is to automatically incorporate information about syntactic neighborhoods of source words into these feature vectors, and, thus, potentially improve quality of the translation output.  Since vectors correspond to words, it is natural for us to use dependency syntax. Dependency trees (see Figure~\ref{fig:deptree}) represent syntactic relations between words: for example, 
 {\it monkey} is a subject of the predicate {\it eats}, and {\it banana} is its object. 

In order to produce syntax-aware feature representations of words, we exploit graph-convolutional networks (GCNs) ~\cite{duvenaud2015convolutional,DefferrardBV16,kearnes2016molecular,kipf2016semigraphconv}. GCNs can be regarded as computing a latent-feature representation of a node (i.e. a real-valued vector)  based on its $k$-th order neighborhood (i.e. nodes at most $k$ hops aways from the node)~\cite{gilmer2017}. They are generally simple and computationally inexpensive.  We use Syntactic GCNs, a version of GCN operating on top of syntactic dependency trees,  recently shown effective in the context of semantic role labeling
\cite{marcheggiani-titov:2017:srlgcn}. 

Since syntactic GCNs produce representations at word level, it is straightforward to use them as encoders within the attention-based encoder-decoder framework. As NMT systems are trained end-to-end,  GCNs end up  capturing syntactic properties specifically relevant to the translation task. Though GCNs can take word embeddings as input,
we will see that they are more effective when used as layers on top of recurrent neural network (RNN) or convolutional neural network (CNN) encoders~\cite{gehring2016convolutional}, enriching their states with syntactic information. A comparison to RNNs is 
the most challenging test for GCNs, as it has been shown that RNNs (e.g., LSTMs)
 are able to capture certain syntactic phenomena (e.g., subject-verb agreement) reasonably well on their own, without explicit treebank supervision~\cite{linzen-dupoux-goldberg:2016:tacllstmsyntax,shi2016emnlp}. Nevertheless, GCNs appear beneficial even in this challenging set-up: 
we obtain +1.2 and +0.7 BLEU point improvements from using syntactic GCNs on top of bidirectional RNNs for English-German and English-Czech, respectively.

In principle, GCNs are flexible enough to incorporate any linguistic structure as long as they can be represented as graphs (e.g., dependency-based semantic-role labeling representations~\cite{surdeanuJMMN08}, AMR semantic graphs~\cite{banarescu2012abstract} and co-reference chains). For example, unlike recursive neural networks~\cite{socher-EtAl:2013:EMNLP}, GCNs do not require the graphs to be trees. However, in this work we solely focus on dependency syntax and leave more general investigation for future work.

Our main contributions can be summarized as follows:
\begin{itemize}
\item we introduce a method for incorporating structure into NMT using syntactic GCNs;
\item we show that GCNs can be used along with RNN and CNN encoders;
\item we show that incorporating structure is beneficial for machine translation on English-Czech and English-German.
\end{itemize}

%% file: background.tex
\section{Background}
\label{sec:background}

\paragraph{Notation.} We use $\vec{x}$ for vectors, $\vec{x}_{1:t}$ for a sequence of $t$ vectors, and
$X$ for matrices. The $i$-th \textit{value} of vector $\mathbf{x}$ is denoted by $x_i$. We use $\circ$ for vector concatenation.

\subsection{Neural Machine Translation}
\label{sec:nmt}

In NMT \cite{kalchbrenner13emnlp,sutskever2014seq2seqnips,cho14emnlp}, given example translation pairs from a parallel corpus, a neural network is trained to directly estimate the conditional distribution $p(y_{1:T_y} | x_{1:T_x})$ of translating a source sentence $x_{1:T_x}$ (a sequence of $T_x$ words) into a target sentence $y_{1:T_y}$. NMT models typically consist of an encoder, a decoder and some method for conditioning the decoder on the encoder, for example, an attention mechanism. We will now briefly describe the components that we use in this paper.

\subsubsection{Encoders}
\label{sec:encoders}
An encoder is a function that takes as input the source sentence and produces a representation encoding its semantic content.
We describe recurrent, convolutional and bag-of-words encoders.

\paragraph{Recurrent.} Recurrent neural networks (RNNs) \cite{elman1990finding} model sequential data. They receive one input vector at each time step and update their hidden state to summarize all inputs up to that point. Given an input sequence
$\mathbf{x}_{1:{T_x}} = \mathbf{x}_1, \mathbf{x}_2, \dots, \mathbf{x}_{T_x} $ of word embeddings
an RNN is defined recursively as follows:
$$\textsc{RNN}(\mathbf{x}_{1:t}) = f(\mathbf{x}_t , \textsc{RNN}(\mathbf{x}_{1:t-1}))$$
\noindent where $f$ is a nonlinear function such as an LSTM \cite{hochreiter97} or a GRU \cite{cho14emnlp}. We will use the function \textsc{RNN} as an abstract mapping from an input sequence $\mathbf{x}_{1:T}$ to final hidden state $\textsc{RNN}(\mathbf{x}_{1:T_x})$, regardless of the used nonlinearity. To not only summarize the past of a word, but also its future, a bidirectional RNN \cite{schusterpaliwal1997,irsoycardie14emnlp} is often used. A bidirectional RNN reads the input sentence in two directions and then concatenates the states for each time step:
$$\textsc{BiRNN}(\mathbf{x}_{1:T_x}, t) = \textsc{RNN}_F(\mathbf{x}_{1:t}) \circ \textsc{RNN}_B(\mathbf{x}_{T_x:t})   $$
\noindent where $\textsc{RNN}_F$ and $\textsc{RNN}_B$ are the forward and backward RNNs, respectively. For further details we refer to  the encoder of \newcite{bahdanau15iclr}.

\paragraph{Convolutional.} Convolutional Neural Networks (CNNs) apply a fixed-size window over the input sequence to capture the local context of each word \cite{gehring2016convolutional}. One advantage of this approach over RNNs is that it allows for fast parallel computation, while sacrificing non-local context. To remedy the loss of context, multiple CNN layers can be stacked. Formally, given an input sequence $\mathbf{x}_{1:T_x}$, we define a CNN as follows:
$$\textsc{CNN}(\vec{x}_{1:T_x}, t) = f(\vec{x}_{t-\lfloor w/2 \rfloor}, .., \vec{x}_t, .., \vec{x}_{t+\lfloor w/2 \rfloor}  )$$ 
\noindent where $f$ is a nonlinear function, typically a linear transformation followed by ReLU, and $w$ is the size of the window. 

\paragraph{Bag-of-Words.} In a bag-of-words (BoW) encoder every word is simply represented by its word embedding. To give the decoder some sense of word position, position embeddings (PE) may be added. There are different strategies for defining position embeddings, and in this paper we choose to learn a vector for each absolute word position up to a certain maximum length. We then represent the $t$-th word in a sequence as follows:
$$\textsc{BoW}(\vec{x}_{1:T_x}, t) = \vec{x}_t + \vec{p}_t$$
\noindent where $\vec{x}_t$ is the word embedding and $\vec{p}_t$ is the t-th position embedding.

\subsubsection{Decoder}
A decoder produces the target sentence conditioned on the representation of the source sentence induced by the encoder. In \newcite{bahdanau15iclr} the decoder is implemented as an \textsc{RNN} conditioned on an additional input $\vec{c}_i$, the context vector, which is dynamically computed at each time step using an attention mechanism. 

The probability of a target word $y_i$ is now a function of the decoder RNN state, the previous target word embedding, and the context vector.
The model is trained end-to-end for maximum log likelihood of the next target word given its context.

\subsection{Graph Convolutional Networks}
\label{sec:gcn}
We will now describe the Graph Convolutional Networks (GCNs) of \newcite{kipf2016semigraphconv}. For a comprehensive overview of alternative GCN architectures see~\newcite{gilmer2017}.

A GCN is a multilayer neural network that operates directly on a graph, encoding information about the neighborhood of a node as a real-valued vector. In each GCN layer, information flows along edges of the graph; in other words, each node receives messages from all its immediate neighbors. When multiple GCN layers are stacked, information about larger neighborhoods gets integrated.
For example, in the second layer, a node will receive information from its immediate neighbors, but this information already includes information from their respective neighbors. 
By choosing the number of GCN layers, we regulate the distance the information travels: with $k$ layers a node receives information from neighbors at most $k$ hops away. 

Formally, consider an undirected graph $\mathcal{G} = (\mathcal{V}, \mathcal{E})$, where $\mathcal{V}$ is a set of $n$ nodes, and $\mathcal{E}$ is a set of edges. Every node is assumed to be connected to itself, i.e. $\forall v\in \mathcal{V} : (v, v) \in \mathcal{E}.$ 
Now, let $X \in \mathbb{R}^{d \times n}$  be a matrix containing all $n$ nodes with their features, where $d$ is the dimensionality of the feature vectors. 
In our case, $X$ will contain word embeddings, but in general it can contain any kind of features. For a 1-layer GCN, the new node representations are computed as follows:
\begin{equation*}
\vec{h}_v = \rho \Bigg(  \sum_{u \in \mathcal{N}(v)} W \vec{x}_u + \vec{b} \Bigg)
\label{eq:gcn-singlelayer-prop-rule}
\end{equation*}
\noindent where $W \in \mathbb{R}^{d \times d}$ is a  weight matrix and $\vec{b} \in \mathbb{R}^d$ a bias vector.\footnote{We dropped the normalization factor used by \newcite{kipf2016semigraphconv}, as it is not used in syntactic GCNs of~\newcite{marcheggiani-titov:2017:srlgcn}.} $\rho$ is an activation function, e.g. a ReLU. $\mathcal{N}(v)$ is the set of neighbors of $v$, which we assume here to always include $v$ itself. 
As stated before, to allow information to flow over multiple hops, we need to stack GCN layers. The recursive computation is as follows:
\begin{equation*}
\vec{h}_v^{(j+1)} = \rho \Bigg(  \sum_{u \in \mathcal{N}(v)} W^{(j)} \vec{h}_u^{(j)} + \vec{b}^{(j)} \Bigg)
\label{eq:gcn-multilayer-prop-rule}
\end{equation*}
\noindent where $j$ indexes the layer, and $\vec{h}_v^{(0)} = \vec{x}_v$.

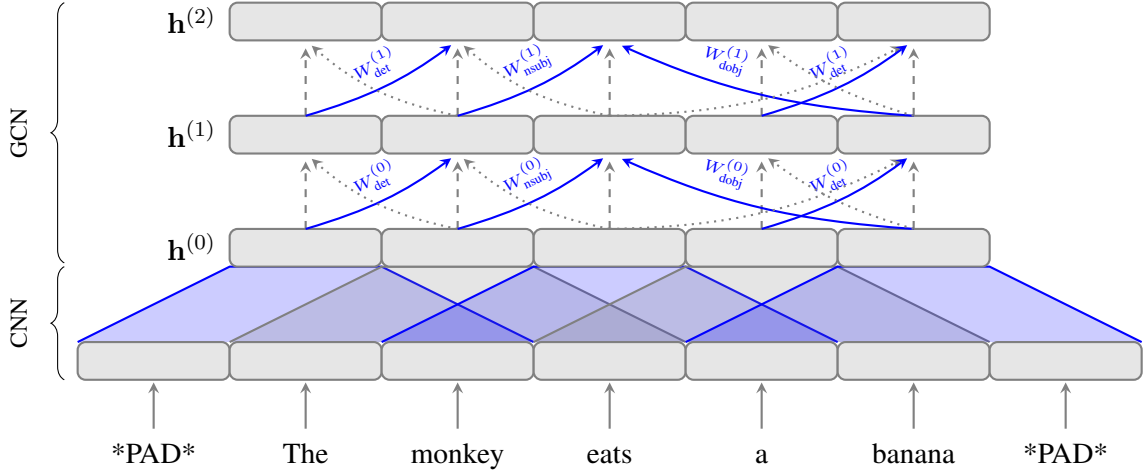
\begin{figure*}[t]
\begin{center}
\input{figures/conv_gcn.tikz}
\end{center}
\vspace{-2ex}
\caption{A 2-layer syntactic GCN on top of a convolutional encoder. Loop connections are depicted with dashed edges, syntactic ones with solid (dependents to heads) and dotted (heads to dependents) edges. Gates and some labels are omitted for clarity.}
\label{fig:conv-gcn}
\end{figure*}

\subsection{Syntactic GCNs}
\label{sec:syngcn}
\newcite{marcheggiani-titov:2017:srlgcn} generalize GCNs to operate on directed and labeled graphs.\footnote{For an alternative approach to integrating labels and directions, see applications of GCNs to statistical relation learning~\cite{schlichtkrull2017}.}
This makes it possible to use linguistic structures such as dependency trees, where directionality and edge labels play an important role. They also integrate edge-wise gates which let the model regulate contributions of individual dependency edges.
We will briefly describe these modifications.

\paragraph{Directionality.} In order to deal with directionality of edges, separate weight matrices are used for incoming and outgoing edges. We follow the convention that in dependency trees heads point to their dependents, and thus \textit{outgoing} edges are used for head-to-dependent connections, and \textit{incoming} edges are used for dependent-to-head connections.
Modifying the recursive computation for directionality, we arrive at:
\begin{equation*}
\vec{h}_v^{(j+1)} = \rho \Bigg(  \sum_{u \in \mathcal{N}(v)} W^{(j)}_{\dir(u, v)} \, \vec{h}_u^{(j)} + \vec{b}^{(j)}_{\dir(u, v)} \Bigg)
\end{equation*}
\noindent where $\dir(u,v)$ selects the weight matrix associated with the directionality of the edge connecting $u$ and $v$ (i.e. $W_{\text{IN}}$ for $u$-to-$v$, $W_{\text{OUT}}$ for $v$-to-$u$, and $W_{\text{LOOP}}$ for $v$-to-$v$). 
Note that self loops are  modeled separately,

so there are now three times as many parameters as in a non-directional GCN.

\paragraph{Labels.} Making the GCN sensitive to labels is straightforward given the above modifications for directionality. Instead of using separate matrices for each direction, separate matrices are now defined for each direction and label combination:
\begin{equation*}
\vec{h}_v^{(j+1)} = \rho \Bigg( \sum_{u \in \mathcal{N}(v)} W^{(j)}_{\lab(u, v)} \, \vec{h}_u^{(j)} + \vec{b}^{(j)}_{\lab(u, v)} \Bigg)
\end{equation*}
\noindent where we incorporate the directionality of an edge directly in its label.

Importantly, to prevent over-parametrization, only bias terms are made label-specific, in other words: $W_{\lab(u, v)} = W_{\dir(u, v)}$. The resulting syntactic GCN is illustrated in Figure~\ref{fig:conv-gcn} (shown on top of a CNN, as we will explain in the subsequent section). 

\paragraph{Edge-wise gating.} 
Syntactic GCNs also include gates, which can down-weight the contribution of individual edges. They also allow the model to deal with noisy predicted structure, i.e. to ignore potentially erroneous syntactic edges. For each edge, a scalar gate is calculated as follows:
\begin{equation*}
g_{u, v}^{(j)}=\sigma \Big( \vec{h}_u^{(j)} \cdot \vec{\hat{w}}_{\dir(u,v)}^{(j)} + \hat{b}^{(j)}_{\lab(u, v)} \Big)
\end{equation*}
\noindent where $\sigma$ is the logistic sigmoid function, and  $\vec{\hat{w}}_{\dir(u,v)}^{(j)} \in \mathbb{R}^d$ and $\hat{b}^{(j)}_{\lab(u, v)} \in \mathbb{R}$ are learned parameters for the gate. The computation becomes:
\begin{align*}
\vec{h}_v^{(j+1)}\!\!=&\rho\Big(\!\!\!\sum_{u \in \mathcal{N}(v)} \!\!\! g_{u, v}^{(j)} \big(W^{(j)}_{\dir(u, v)} \, \vec{h}_u^{(j)} + \vec{b}^{(j)}_{\lab(u, v)}\big)\Big)
\end{align*}

%% file: figures/conv_gcn.tikz
\usetikzlibrary{decorations.pathreplacing}
\begin{tikzpicture}[scale=0.5,node distance=1cm]

\draw [fill=blue,draw=blue,thick,fill opacity=0.2]  (0*4,1) --  (0*4+4,3) --  (0*4+8,3) -- (0*4+12,1);
\draw [fill=gray,draw=gray,thick,fill opacity=0.2]  (1*4,1) --  (1*4+4,3) --  (1*4+8,3) -- (1*4+12,1);
\draw [fill=blue,draw=blue,thick,fill opacity=0.2]  (2*4,1) --  (2*4+4,3) --  (2*4+8,3) -- (2*4+12,1);
\draw [fill=gray,draw=gray,thick,fill opacity=0.2]  (3*4,1) --  (3*4+4,3) --  (3*4+8,3) -- (3*4+12,1);
\draw [fill=blue,draw=blue,thick,fill opacity=0.2]  (4*4,1) --  (4*4+4,3) --  (4*4+8,3) -- (4*4+12,1);

\tikzstyle{vector}=[thick,draw=black!50,fill=black!10,rounded corners=3pt]

\foreach \x [count = \xi] in {0,...,6} { 
\filldraw[vector] (\x*4,0) rectangle (\x*4+4,1);
}

\foreach \x [count = \xi] in {1,...,5} { 
  \filldraw[vector] (\x*4,3) rectangle (\x*4+4,4);
}

\foreach \x [count = \xi] in {1,...,5} { 
  \filldraw[vector] (\x*4,6) rectangle (\x*4+4,7);
}

\foreach \x [count = \xi] in {1,...,5} { 
  \filldraw[vector] (\x*4,9) rectangle (\x*4+4,10);
}

\foreach \x [count = \xi] in {1,...,5} { 
  \coordinate (GCN0-M\x) at (\x*4+2,4); 
  \coordinate (GCN0-T\x) at (\x*4+2,6); 
  \coordinate (GCN1-M\x) at (\x*4+2,7); 
  \coordinate (GCN1-T\x) at (\x*4+2,9); 

  \coordinate (TOP-M\x) at (\x*4+2,10); 
  \coordinate (TOP-T\x) at (\x*4+2,11); 

}

\foreach \x [count = \xi] in {1,...,5} { 
  \draw[gray,thick,dashed,->,>=stealth,shorten >=3pt,auto] (GCN0-M\x) to (GCN0-T\x);
  \draw[gray,thick,dashed,->,>=stealth,shorten >=3pt,auto] (GCN1-M\x) to (GCN1-T\x);
}

\foreach \y [count = \yi] in {0,...,1} { 
  \path[thick,blue,->,>=stealth,shorten >=3pt,auto,bend right=10] (GCN\y-M1) edge [left,font=\tiny] node [pos=0.5, sloped, above] {$W_{\text{det}}^{(\y)}$} (GCN\y-T2);
  \path[gray,thick,dotted,->,>=stealth,shorten >=3pt,auto,bend left=15] (GCN\y-M2) edge [left,font=\tiny] node [pos=0.6, sloped, above] {} (GCN\y-T1);
  
  \path[thick,blue,->,>=stealth,shorten >=3pt,auto,bend right=10] (GCN\y-M2) edge [left,font=\tiny] node [pos=0.5, sloped, above] {$W_{\text{nsubj}}^{(\y)}$} (GCN\y-T3);
   \path[gray,thick,dotted,->,>=stealth,shorten >=3pt,auto,bend left=15] (GCN\y-M3) edge [left,font=\tiny] node [pos=0.6, sloped, above] {} (GCN\y-T2); 
  
  \path[thick,blue,->,>=stealth,shorten >=5pt,auto,bend left=10] (GCN\y-M5) edge [right,font=\tiny] node [pos=0.64, sloped, above] {$W_{\text{dobj}}^{(\y)}$} (GCN\y-T3);
    \path[gray,thick,dotted,->,>=stealth,shorten >=5pt,auto,bend right=13.5] (GCN\y-M3) edge [right,font=\tiny] node [pos=0.6, sloped, above] {} (GCN\y-T5);
  
  \path[thick,blue,->,>=stealth,shorten >=3pt,auto,bend right=10] (GCN\y-M4) edge [left,font=\tiny] node [pos=0.5, sloped, above] {$W_{\text{det}}^{(\y)}$} (GCN\y-T5);
  \path[gray,thick,dotted,->,>=stealth,shorten >=3pt,auto,bend left=10] (GCN\y-M5) edge [left,font=\tiny] node [pos=0.6, sloped, above] {} (GCN\y-T4);  
}

\foreach \x [count = \xi] in {0,...,6} { 
\draw[gray,thick,solid,->,>=stealth,shorten >=1pt,auto,shorten <=10pt] (\x*4+2,-2) to (\x*4+2,0); 
}

\tikzstyle{word}=[align=center,minimum height=3em,
    text height=1.5ex,
    text depth=.25ex,
    text width=11em,
    text centered,
    minimum height=3em,]
\node[word] at (2,-2) {*PAD*};
\node[word] at (6,-2) {The};
\node[word] at (10,-2) {monkey};
\node[word] at (14,-2) {eats};
\node[word] at (18,-2) {a};
\node[word] at (22,-2) {banana};
\node[word] at (26,-2) {*PAD*};

\node[word] at (3,3.4) {$\vec{h}^{(0)}$};
\node[word] at (3,6.4) {$\vec{h}^{(1)}$};
\node[word] at (3,9.4) {$\vec{h}^{(2)}$};


\draw [decorate,decoration={brace,amplitude=5pt},xshift=-10pt,yshift=0pt]
(0.0,3.1) -- (0.0,10.0) node [black,midway,xshift=-0.6cm,rotate=90] 
{\footnotesize GCN};

\draw [decorate,decoration={brace,amplitude=5pt},xshift=-10pt,yshift=0pt]
(0.0,0.0) -- (0.0,3.0) node [black,midway,xshift=-0.6cm,rotate=90] 
{\footnotesize CNN};


\end{tikzpicture}

%% file: model.tex
\section{Graph Convolutional Encoders}
\label{sec:model}

In this work we focus on exploiting structural information on the source side, i.e. in the encoder. We hypothesize that using an encoder that incorporates syntax will lead to more informative representations
of words, and that these representations, when used as context vectors by the decoder, will lead to an improvement in translation quality.
Consequently, in all our models, we use the decoder of \newcite{bahdanau15iclr} and keep this part of the model constant. As is now common practice, we do not use a maxout layer in the decoder, but apart from this we do not deviate from the original definition. In all models we make use of GRUs \cite{cho14emnlp} as our RNN units.

Our models vary in the encoder part, where we exploit the power of GCNs to induce syntactically-aware representations. We now define a series of  encoders of increasing complexity.

\paragraph{BoW + GCN.} In our first and simplest model, we propose a bag-of-words encoder (with position embeddings, see \S\ref{sec:encoders}), with a GCN on top. In other words, inputs $\mathbf{h}^{(0)}$ are a sum of embeddings of a word and its position in a sentence. Since the original BoW encoder captures the linear ordering information only in a very crude way (through the position embeddings), the structural information provided by GCN
should be highly beneficial.

\paragraph{Convolutional + GCN.} In our second model, we use convolutional neural networks to learn word representations. CNNs are fast, but by definition only use a limited window of context. Instead of the approach used by \newcite{gehring2016convolutional} (i.e. stacking multiple CNN layers on top of each other), we use a GCN to enrich the one-layer CNN representations. Figure~\ref{fig:conv-gcn} shows this model. Note that, while the figure shows a CNN with a window size of 3, we will use a larger window size of 5 in our experiments. We expect this model to perform better than BoW + GCN, because of the additional local context captured by the CNN. 

\paragraph{BiRNN + GCN.} In our third and most powerful model, we employ bidirectional recurrent neural networks. 
In this model, we start by encoding the source sentence using a BiRNN (i.e. BiGRU), and use the resulting hidden states as input to a GCN.  Instead of relying on linear order only, the GCN will allow the encoder to `teleport' over parts of the input sentence, along dependency edges, connecting words that otherwise might be far apart. The model might not only benefit from this teleporting capability however; also the nature of the relations between words (i.e. dependency relation types) may be useful, and the GCN exploits this information (see \S\ref{sec:syngcn} for details).

This is the most challenging setup for GCNs, as RNNs have been shown capable of capturing at least some degree of syntactic information without explicit supervision~\cite{linzen-dupoux-goldberg:2016:tacllstmsyntax}, and hence they should be hard to improve on by incorporating treebank syntax. 

\newcite{marcheggiani-titov:2017:srlgcn} did not observe improvements from using multiple GCN layers in semantic role labeling. However, we do expect that propagating information from further in the tree should be beneficial in principle.
We hypothesize that the first layer is the most influential one, capturing most of the syntactic context, and that additional layers only modestly modify the representations. To ease optimization, we add a residual connection \cite{he2016deep} between the GCN layers, when using more than one layer.

%% file: experiments.tex
\section{Experiments}
\label{sec:experiments}

Experiments are performed using the Neural Monkey toolkit\footnote{https://github.com/ufal/neuralmonkey} \cite{NeuralMonkey:2017}, which implements the model of \newcite{bahdanau15iclr} in TensorFlow. We use the Adam optimizer \cite{kingma2015adam} with a learning rate of 0.001 (0.0002 for CNN models).\footnote{Like \newcite{gehring2016convolutional} we note that Adam is too aggressive for CNN models, hence we use a lower learning rate.} The batch size is set to 80. Between layers we apply dropout with a probability of 0.2, and in experiments with GCNs\footnote{GCN code at https://github.com/bastings/neuralmonkey} we use the same value for edge dropout. We train for 45 epochs, evaluating the BLEU performance of the model every epoch on the validation set. For testing, we select the model with the highest validation BLEU. L2 regularization is used with a value of $10^{-8}$. All the model selection (incl. hyperparameter selections) was performed on the validation set.
In all experiments we obtain translations using a greedy decoder, i.e. we select the output token with the highest probability at each time step. 

We will describe an artificial  experiment in \S\ref{sec:artificial} and MT experiments in \S\ref{sec:news-experiments}. 

\subsection{Reordering artificial sequences}
\label{sec:artificial}
Our goal here is to provide an intuition for the capabilities of GCNs.
We define a reordering task where randomly permuted sequences need to be put back into the original order. We encode the original order using edges, and test if GCNs can successfully exploit them. Note that this task is not meant to provide a fair comparison to RNNs. The input (besides the edges) simply does not carry any information about the original ordering, so RNNs cannot possibly solve this task.

\paragraph{Data.} From a vocabulary of 26 types,
we generate random sequences of 3-10 tokens. We then randomly permute them, pointing every token to its original predecessor with a label sampled from a set of 5 labels. Additionally, we  point every token to an \textit{arbitrary} position in the sequence with a label from a distinct set of 5 `fake' labels. We sample 25000 training and 1000 validation sequences.

\paragraph{Model.} We use the BiRNN + GCN model, i.e. a bidirectional GRU with a 1-layer GCN on top. We use 32, 64 and 128 units for embeddings, GRU units and GCN layers, respectively. 

\begin{figure}[h]
\centering
\includegraphics[width=8cm,trim=5 0 5 40,clip]{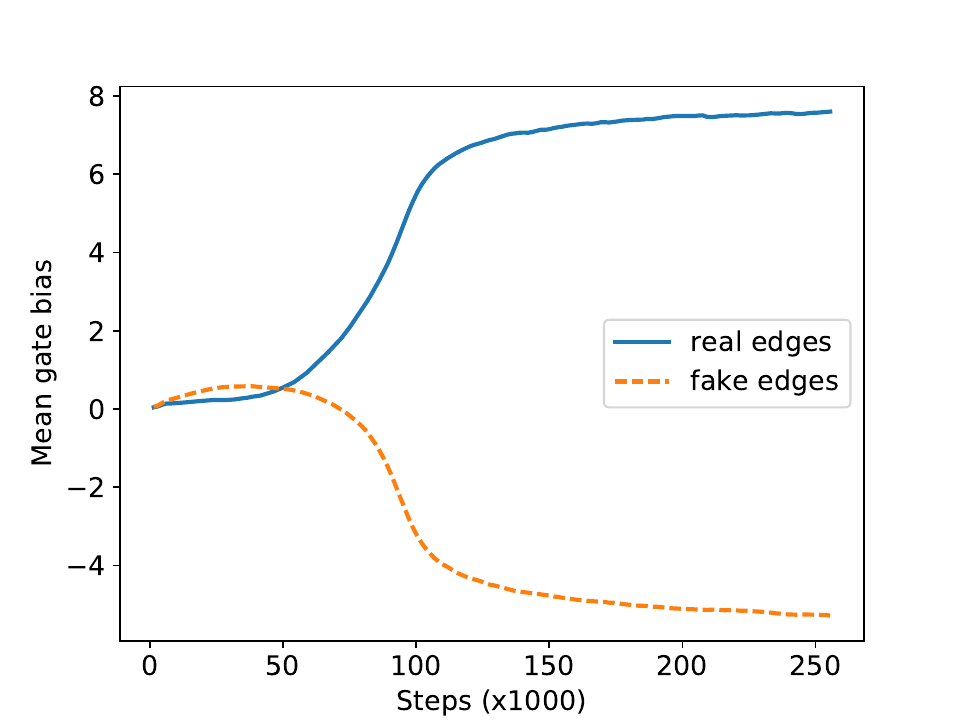}
\caption{Mean values of gate bias terms for real (useful) labels and for fake (non useful) labels suggest the GCN learns to distinguish them.}
\label{fig:artificial}
\end{figure}

\paragraph{Results.} After 6 epochs of training, the model learns to put permuted sequences back into order, reaching a validation BLEU of $99.2$. Figure~\ref{fig:artificial} shows that the mean values of the bias terms of gates (i.e. $\hat{b}$) for real and fake edges are far apart, suggesting that the GCN learns to distinguish them. Interestingly, this illustrates why edge-wise gating is beneficial. A gate-less model would not understand which of the two outgoing arcs is fake and which is genuine, because only biases $b$ would then be label-dependent. Consequently, it would only do a mediocre job in reordering. Although
using label-specific matrices $W$ would also 
help, this would not scale to the real scenario (see \S\ref{sec:syngcn}).

\subsection{Machine Translation}
\label{sec:news-experiments}

\paragraph{Data.} For our experiments we use the  En-De and En-Cs News Commentary v11 data from the WMT16 translation task.\footnote{http://www.statmt.org/wmt16/translation-task.html} For En-De we also train on the full  WMT16 data set.
As our validation set and test set we use \texttt{newstest2015} and \texttt{newstest2016}, respectively.

\paragraph{Pre-processing.} The English sides of the corpora are tokenized and parsed into dependency trees by SyntaxNet,\footnote{https://github.com/tensorflow/models/tree/master/syntaxnet} using the pre-trained Parsey McParseface model.\footnote{The used dependency parses can be reproduced by using the \texttt{syntaxnet/demo.sh} shell script.}  The Czech and German sides are tokenized using the Moses tokenizer.\footnote{https://github.com/moses-smt/mosesdecoder} Sentence pairs where either side is longer than 50 words are filtered out after tokenization.

\paragraph{Vocabularies.} For the English sides, we construct vocabularies from all words except those with a training set frequency smaller than three. For Czech and German, to deal with rare words and phenomena such as inflection and compounding, we learn byte-pair encodings (BPE) as described by \newcite{sennrich2016subword}. Given the size of our data set, and following \newcite{google2016nmt}, we use 8000 BPE merges to obtain robust frequencies for our subword units (16000 merges for full data experiment). Data set statistics are summarized in Table~\ref{tab:data} and vocabulary sizes in Table~\ref{tab:voc}. 

\begin{table}[bh]
\centering
\begin{tabular}{@{}lrrr@{}} 
\toprule
          & Train    & Val. & Test \\ 
\midrule
English-German & 226822 & 2169 & 2999 \\
English-German (full) & 4500966  & 2169 & 2999\\
English-Czech & 181112 & 2656 & 2999 \\
\bottomrule
\end{tabular}
\caption{The number of sentences in our data sets.}
\label{tab:data}
\end{table}

\begin{table}[th]
\centering
\begin{tabular}{@{}lrr@{}} 
\toprule
           & Source    & Target \\ 
\midrule
English-German & 37824 & 8099 (BPE) \\
English-German (full)      & 50000 & 16000 (BPE) \\
English-Czech  & 33786 & 8116 (BPE) \\
\bottomrule
\end{tabular}
\caption{Vocabulary sizes.}
\label{tab:voc}
\end{table}

\paragraph{Hyperparameters.} We use 256 units for word embeddings, 512 units for GRUs (800 for En-De full data set experiment), and 512 units for convolutional layers (or equivalently, 512 `channels'). The dimensionality of the GCN layers is equivalent to the dimensionality of their input. We report results for 2-layer GCNs, as we find them most effective (see ablation studies below).

\paragraph{Baselines.} We provide three baselines, each with a different encoder: a bag-of-words encoder, a convolutional encoder with window size $w=5$, and a BiRNN. See \S\ref{sec:encoders} for details. 

\paragraph{Evaluation.} We report (cased) BLEU results \cite{papineni2002bleu} using \texttt{multi-bleu}, as well as Kendall $\tau$ reordering scores.\footnote{See \newcite{stanojevic2015evaluating}. TER \cite{snover2006ter} and BEER \cite{stanojevic-simaan:2014:EMNLP2014} metrics, even though omitted due to space considerations, are consistent with the reported results.} 

\subsubsection{Results} 
\paragraph{English-German.} Table~\ref{tab:test-en2de} shows test results on English-German. 
Unsurprisingly, the bag-of-words baseline performs the worst.  
We expected the BoW+GCN model to make easy gains over this baseline, which is indeed what happens.
The CNN baseline reaches a higher \bleu score than the BoW models, but interestingly its \bleuone score is lower than the BoW+GCN model. The CNN+GCN model improves over the CNN baseline by +1.9 and +1.1 for \bleuone and \bleu, respectively. The BiRNN, the strongest baseline, reaches a \bleu of 14.9. Interestingly, GCNs still manage to improve the result by +2.3 \bleuone and +1.2 \bleu points. 
Finally, we observe a big jump in \bleu by using the full data set and beam search (beam 12). The BiRNN now reaches 23.3, while adding a GCN achieves a score of 23.9.

\input{table_test_en2de.tex}

\paragraph{English-Czech.} Table~\ref{tab:test-en2cs} shows test results on English-Czech. While it is difficult to obtain high absolute BLEU scores on this dataset, we can still see similar relative improvements. Again the BoW baseline scores worst, with the BoW+GCN easily beating that result.
The CNN baseline scores \bleu of 8.1, but the CNN+GCN improves on that, this time by +1.0 and +0.6 for \bleuone and \bleu, respectively. Interestingly, \bleuone scores for the BoW+GCN and CNN+GCN models are higher than both baselines so far. Finally, the BiRNN baseline scores a \bleu of 8.9, but it is again beaten by the BiRNN+GCN model with +1.9 \bleuone and +0.7 \bleu.

\input{table_test_en2cs.tex}

\paragraph{Effect of GCN layers.} How many GCN layers do we need? Every layer gives us an extra hop in the graph and expands the syntactic neighborhood of a word. Table~\ref{tab:gcn-layers} shows validation BLEU performance as a function of the number of GCN layers. For English-German, using a 1-layer GCN improves BLEU-1, but surprisingly has little effect on \bleu. Adding an additional layer gives improvements on both \bleuone and \bleu of +1.3 and +0.73, respectively. For English-Czech, performance increases with each added GCN layer.

\begin{table}[h]
\centering
\begin{tabular}{@{}lrrrr@{}} \toprule
 & \multicolumn{2}{c}{En-De} & \multicolumn{2}{c}{En-Cs}\\ 
     & \footnotesize{\bleuone}&\footnotesize{\bleu} & \footnotesize{\bleuone}&\footnotesize{\bleu}\\
\midrule
BiRNN			 & 44.2 & 14.1 & 37.8 & 8.9\\  
\, + GCN (1L)    & 45.0 & 14.1 & 38.3 & 9.6\\
\, + GCN (2L)    & 46.3 & 14.8 & 39.6 & 9.9\\
\bottomrule
\end{tabular}
\caption{Validation BLEU for English-German and English-Czech for 1- and 2-layer GCNs.}
\label{tab:gcn-layers}
\end{table}

\begin{figure}[h]
\centering
\includegraphics[width=\columnwidth,trim=5 0 5 40,clip]{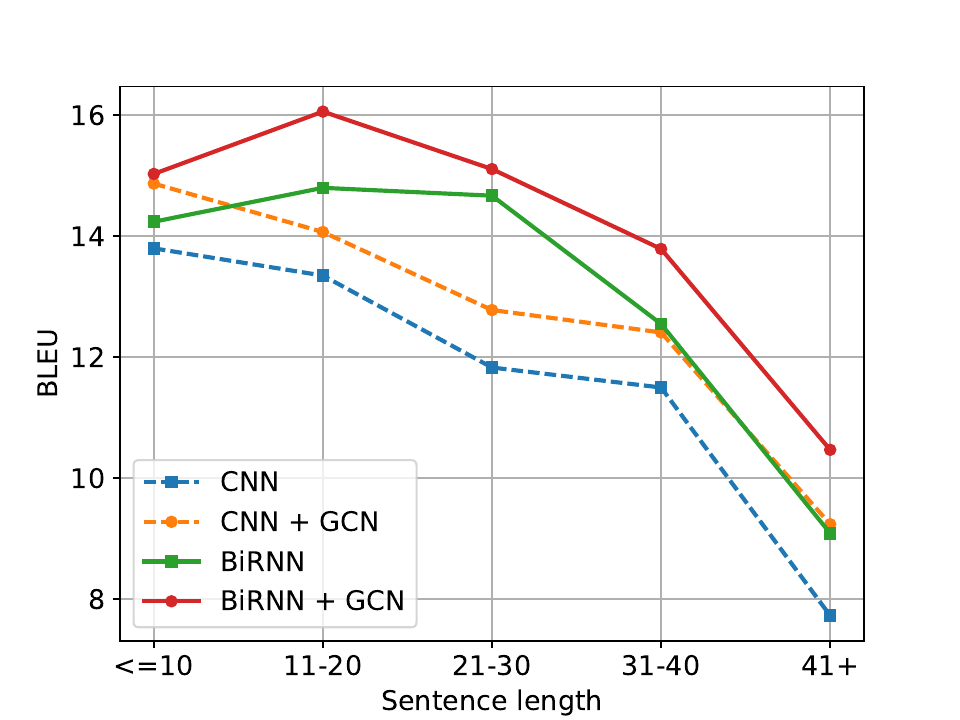}
\caption{Validation BLEU per sentence length. 
}
\label{fig:bleu-per-sentence}
\end{figure}

\paragraph{Effect of sentence length.} 
We  hypothesize that GCNs should be more beneficial for longer sentences: these are likely to contain long-distance syntactic dependencies which may not be adequately captured by RNNs but directly encoded in GCNs. To test this, we partition the validation data into five buckets and calculate BLEU for each of them. Figure~\ref{fig:bleu-per-sentence} shows that GCN-based models outperform their respective baselines rather uniformly across all buckets. This is a surprising result. One explanation may be that syntactic parses are noisier for longer sentences, and this prevents us from obtaining extra improvements with GCNs.  

\paragraph{Discussion.} 
Results suggest that the syntax-aware representations provided by GCNs consistently lead to improved translation performance as measured by \bleu (as well as TER and BEER). 
Consistent gains in terms of Kendall tau and \bleuone indicate that improvements correlate with better word order and lexical/BPE selection, two phenomena that depend crucially on syntax. 

%% file: table_test_en2de.tex
\begin{table}[h]
\centering
\begin{tabular}{@{}lrrr@{}} \toprule
            & Kendall & \bleuone & \bleu \\
\midrule
BoW         & 0.3352 & 40.6 &  9.5\\
\quad + GCN   & 0.3520 & 44.9 & 12.2\\
\midrule
CNN         & 0.3601 & 42.8 & 12.6\\
\quad + GCN   & 0.3777 & 44.7 & 13.7\\
\midrule
BiRNN       & 0.3984 & 45.2 & 14.9\\
\quad + GCN & 0.4089 & 47.5 & 16.1\\
\midrule
BiRNN (full)       & 0.5440 &  53.0 & 23.3\\
 \quad + GCN & 0.5555 &  54.6 & 23.9\\
\bottomrule
\end{tabular}
\caption{Test results for English-German.}
\label{tab:test-en2de}
\end{table}

%% file: table_test_en2cs.tex
\begin{table}[h]
\centering
\begin{tabular}{@{}lrrr@{}} \toprule
           & Kendall & \bleuone & \bleu\\
\midrule
BoW         & 0.2498 & 32.9 & 6.0 \\
\quad + GCN   & 0.2561 & 35.4 & 7.5\\
\midrule
CNN         & 0.2756 & 35.1 & 8.1\\
\quad + GCN   & 0.2850 & 36.1 & 8.7\\
\midrule
BiRNN       & 0.2961 & 36.9 & 8.9\\
\quad + GCN & 0.3046 & 38.8 & 9.6\\
\bottomrule
\end{tabular}
\caption{Test results for English-Czech.}
\label{tab:test-en2cs}
\end{table}

%% file: relatedwork.tex
\section{Related Work}
\label{sec:relatedwork}

We review various accounts to syntax in NMT as well as other convolutional encoders.

\paragraph{Syntactic features and/or constraints.} 
\newcite{sennrich2016linguistic} embed features such as POS-tags, lemmas and dependency labels and feed these into the network along with word embeddings.
\newcite{eriguchi2016treetoseq} parse English sentences with an HPSG parser and use a Tree-LSTM to encode the internal nodes of the tree. In the decoder, word and node representations compete under the same attention mechanism. 
\newcite{stahlberg2016syntactically} use a pruned lattice from a hierarchical phrase-based model (hiero) to constrain NMT.
Hiero trees are not syntactically-aware, but instead constrained by symmetrized word alignments. \newcite{aharonigoldberg2017stringtotree} propose neural string-to-tree by predicting linearized parse trees.\looseness=-1

\paragraph{Multi-task Learning.}
Sharing NMT parameters with a syntactic parser is a popular approach to obtaining syntactically-aware representations.
\newcite{luong16iclr} predict linearized constituency parses as an additional task.
\newcite{eriguchi2017nmtrg} multi-task with a  target-side RNNG parser \cite{dyer-EtAl:2016:RNNG} and improve on various language pairs with English on the target side. \newcite{nadejdeetal2017syntaxaware} multi-task with CCG tagging, and also integrate syntax on the target side by predicting a sequence of words interleaved with CCG supertags.

\paragraph{Latent structure.}
\newcite{hashimoto2017latentgraph} add a syntax-inspired encoder on top of a BiLSTM layer. They encode source words as a learned average of potential parents emulating a relaxed dependency tree. 
While their model is trained purely on translation data, they also experiment with pre-training the encoder using treebank annotation 
and report modest improvements on English-Japanese.
\newcite{YogatamaBDGL16} introduce a model for language understanding and generation that composes words into sentences by inducing unlabeled binary bracketing trees.

\paragraph{Convolutional encoders.}   
\newcite{gehring2016convolutional}
show that CNNs can be competitive to BiRNNs when used as encoders. To increase the receptive field of a word's context they stack multiple CNN layers. \newcite{kalchbrenner2016linear} use convolution in both the encoder and the decoder; they make use of dilation to increase the receptive field.  In contrast to both approaches, we use a GCN informed by dependency structure to increase it. Finally, \newcite{cho14ssst} propose a recursive convolutional neural network which builds a tree out of the word leaf nodes, but which ends up compressing the source sentence in a single vector.

%% file: conclusion.tex
\section{Conclusions}
\label{sec:conclusions}

We have presented a simple and effective approach to integrating syntax into neural machine translation models and have shown consistent \bleu improvements for two challenging language pairs: English-German and English-Czech.
Since GCNs are capable of encoding any kind of graph-based structure, in future work we would like to go beyond syntax, by using semantic annotations such as SRL and AMR, and co-reference chains.

%% file: ack.tex
\section*{Acknowledgments}
We would like to thank Michael Schlichtkrull and Thomas Kipf for their suggestions and comments.
This work was supported by the European Research Council (ERC StG BroadSem 678254) and the Dutch National Science Foundation (NWO VIDI 639.022.518, NWO VICI 277-89-002).

%% file: ms.bbl
\begin{thebibliography}{47}
\expandafter\ifx\csname natexlab\endcsname\relax\def\natexlab#1{#1}\fi

\bibitem[{{Aharoni} and {Goldberg}(2017)}]{aharonigoldberg2017stringtotree}
Roee {Aharoni} and Yoav {Goldberg}. 2017.
\newblock \href {http://arxiv.org/abs/1704.04743} {{Towards String-to-Tree
  Neural Machine Translation}}.
\newblock \emph{ArXiv e-prints}.

\bibitem[{Bahdanau et~al.(2015)Bahdanau, Cho, and Bengio}]{bahdanau15iclr}
Dzmitry Bahdanau, Kyunghyun Cho, and Yoshua Bengio. 2015.
\newblock \href {http://arxiv.org/abs/1409.0473} {{Neural Machine Translation
  by Jointly Learning to Align and Translate}}.
\newblock In \emph{{Proceedings of the International Conference on Learning
  Representations (ICLR)}}.

\bibitem[{Banarescu et~al.(2012)Banarescu, Bonial, Cai, Georgescu, Griffitt,
  Hermjakob, Knight, Koehn, Palmer, and Schneider}]{banarescu2012abstract}
Laura Banarescu, Claire Bonial, Shu Cai, Madalina Georgescu, Kira Griffitt, Ulf
  Hermjakob, Kevin Knight, Philipp Koehn, Martha Palmer, and Nathan Schneider.
  2012.
\newblock Abstract meaning representation (amr) 1.0 specification.
\newblock In \emph{Conference on Empirical Methods in Natural Language
  Processing}, pages 1533--1544.

\bibitem[{Chiang(2010)}]{chiang:2010:soft}
David Chiang. 2010.
\newblock \href {http://www.aclweb.org/anthology/P10-1146} {Learning to
  translate with source and target syntax}.
\newblock In \emph{Proceedings of the 48th Annual Meeting of the Association
  for Computational Linguistics}, pages 1443--1452, Uppsala, Sweden.
  Association for Computational Linguistics.

\bibitem[{Cho et~al.(2014{\natexlab{a}})Cho, van Merrienboer, Bahdanau, and
  Bengio}]{cho14ssst}
KyungHyun Cho, Bart van Merrienboer, Dzmitry Bahdanau, and Yoshua Bengio.
  2014{\natexlab{a}}.
\newblock \href {http://arxiv.org/abs/1409.1259} {{On the Properties of Neural
  Machine Translation: Encoder-Decoder Approaches}}.
\newblock In \emph{{SSST-8, Eighth Workshop on Syntax, Semantics and Structure
  in Statistical Translation}}, volume abs/1409.1259, pages 103--111.

\bibitem[{Cho et~al.(2014{\natexlab{b}})Cho, van Merrienboer, Gulcehre,
  Bahdanau, Bougares, Schwenk, and Bengio}]{cho14emnlp}
Kyunghyun Cho, Bart van Merrienboer, Caglar Gulcehre, Dzmitry Bahdanau, Fethi
  Bougares, Holger Schwenk, and Yoshua Bengio. 2014{\natexlab{b}}.
\newblock \href {http://www.aclweb.org/anthology/D14-1179} {{Learning Phrase
  Representations using RNN Encoder--Decoder for Statistical Machine
  Translation}}.
\newblock In \emph{{Proceedings of the 2014 Conference on Empirical Methods in
  Natural Language Processing (EMNLP)}}, pages 1724--1734, Doha, Qatar.
  Association for Computational Linguistics.

\bibitem[{Defferrard et~al.(2016)Defferrard, Bresson, and
  Vandergheynst}]{DefferrardBV16}
Micha{\"{e}}l Defferrard, Xavier Bresson, and Pierre Vandergheynst. 2016.
\newblock \href
  {http://papers.nips.cc/paper/6081-convolutional-neural-networks-on-graphs-with-fast-localized-spectral-filtering}
  {Convolutional neural networks on graphs with fast localized spectral
  filtering}.
\newblock In \emph{Advances in Neural Information Processing Systems 29: Annual
  Conference on Neural Information Processing Systems 2016, December 5-10,
  2016, Barcelona, Spain}, pages 3837--3845.

\bibitem[{Duvenaud et~al.(2015)Duvenaud, Maclaurin, Iparraguirre, Bombarell,
  Hirzel, Aspuru-Guzik, and Adams}]{duvenaud2015convolutional}
David~K Duvenaud, Dougal Maclaurin, Jorge Iparraguirre, Rafael Bombarell,
  Timothy Hirzel, Al{\'a}n Aspuru-Guzik, and Ryan~P Adams. 2015.
\newblock Convolutional networks on graphs for learning molecular fingerprints.
\newblock In \emph{Advances in neural information processing systems}, pages
  2224--2232.

\bibitem[{Dyer et~al.(2016)Dyer, Kuncoro, Ballesteros, and
  Smith}]{dyer-EtAl:2016:RNNG}
Chris Dyer, Adhiguna Kuncoro, Miguel Ballesteros, and Noah~A. Smith. 2016.
\newblock \href {http://www.aclweb.org/anthology/N16-1024} {Recurrent neural
  network grammars}.
\newblock In \emph{Proceedings of the 2016 Conference of the North American
  Chapter of the Association for Computational Linguistics: Human Language
  Technologies}, pages 199--209, San Diego, California. Association for
  Computational Linguistics.

\bibitem[{Elman(1990)}]{elman1990finding}
Jeffrey~L Elman. 1990.
\newblock {Finding structure in time}.
\newblock \emph{Cognitive science}, 14(2):179--211.

\bibitem[{Eriguchi et~al.(2016)Eriguchi, Hashimoto, and
  Tsuruoka}]{eriguchi2016treetoseq}
Akiko Eriguchi, Kazuma Hashimoto, and Yoshimasa Tsuruoka. 2016.
\newblock \href {http://www.aclweb.org/anthology/P16-1078} {Tree-to-sequence
  attentional neural machine translation}.
\newblock In \emph{Proceedings of the 54th Annual Meeting of the Association
  for Computational Linguistics (Volume 1: Long Papers)}, pages 823--833,
  Berlin, Germany. Association for Computational Linguistics.

\bibitem[{{Eriguchi} et~al.(2017){Eriguchi}, {Tsuruoka}, and
  {Cho}}]{eriguchi2017nmtrg}
Akiko {Eriguchi}, Yoshimasa {Tsuruoka}, and Kyunghyun {Cho}. 2017.
\newblock \href {http://arxiv.org/abs/1702.03525} {{Learning to Parse and
  Translate Improves Neural Machine Translation}}.
\newblock \emph{ArXiv e-prints}.

\bibitem[{Gehring et~al.(2016)Gehring, Auli, Grangier, and
  Dauphin}]{gehring2016convolutional}
Jonas Gehring, Michael Auli, David Grangier, and Yann~N. Dauphin. 2016.
\newblock \href {http://arxiv.org/abs/1611.02344} {A convolutional encoder
  model for neural machine translation}.
\newblock \emph{CoRR}, abs/1611.02344.

\bibitem[{{Gilmer} et~al.(2017){Gilmer}, {Schoenholz}, {Riley}, {Vinyals}, and
  {Dahl}}]{gilmer2017}
Justin {Gilmer}, Samuel~S. {Schoenholz}, Patrick~F. {Riley}, Oriol {Vinyals},
  and George~E. {Dahl}. 2017.
\newblock \href {http://arxiv.org/abs/1704.01212} {{Neural Message Passing for
  Quantum Chemistry}}.
\newblock \emph{ArXiv e-prints}.

\bibitem[{Hashimoto and Tsuruoka(2017)}]{hashimoto2017latentgraph}
Kazuma Hashimoto and Yoshimasa Tsuruoka. 2017.
\newblock \href {http://arxiv.org/abs/1702.02265} {Neural machine translation
  with source-side latent graph parsing}.
\newblock \emph{CoRR}, abs/1702.02265.

\bibitem[{He et~al.(2016)He, Zhang, Ren, and Sun}]{he2016deep}
Kaiming He, Xiangyu Zhang, Shaoqing Ren, and Jian Sun. 2016.
\newblock Deep residual learning for image recognition.
\newblock In \emph{Proceedings of the IEEE Conference on Computer Vision and
  Pattern Recognition}, pages 770--778.

\bibitem[{Helcl and Libovick{\'{y}}(2017)}]{NeuralMonkey:2017}
Jind{\v{r}}ich Helcl and Jind{\v{r}}ich Libovick{\'{y}}. 2017.
\newblock \href {https://doi.org/10.1515/pralin-2017-0001} {Neural monkey: An
  open-source tool for sequence learning}.
\newblock \emph{The Prague Bulletin of Mathematical Linguistics}, (107):5--17.

\bibitem[{Hochreiter and Schmidhuber(1997)}]{hochreiter97}
Sepp Hochreiter and J{\"u}rgen Schmidhuber. 1997.
\newblock \href {https://doi.org/10.1162/neco.1997.9.8.1735} {{Long Short-Term
  Memory}}.
\newblock \emph{Neural Computation}, 9(8):1735--1780.

\bibitem[{Irsoy and Cardie(2014)}]{irsoycardie14emnlp}
Ozan Irsoy and Claire Cardie. 2014.
\newblock \href {http://www.aclweb.org/anthology/D14-1080} {{Opinion Mining
  with Deep Recurrent Neural Networks}}.
\newblock In \emph{{Proceedings of the 2014 Conference on Empirical Methods in
  Natural Language Processing (EMNLP)}}, pages 720--728, Doha, Qatar.
  Association for Computational Linguistics.

\bibitem[{Kalchbrenner and Blunsom(2013)}]{kalchbrenner13emnlp}
Nal Kalchbrenner and Phil Blunsom. 2013.
\newblock \href {http://www.aclweb.org/anthology/D13-1176} {{Recurrent
  Continuous Translation Models}}.
\newblock In \emph{{Proceedings of the 2013 Conference on Empirical Methods in
  Natural Language Processing}}, pages 1700--1709, Seattle, Washington, USA.

\bibitem[{Kalchbrenner et~al.(2016)Kalchbrenner, Espeholt, Simonyan, van~den
  Oord, Graves, and Kavukcuoglu}]{kalchbrenner2016linear}
Nal Kalchbrenner, Lasse Espeholt, Karen Simonyan, A{\"{a}}ron van~den Oord,
  Alex Graves, and Koray Kavukcuoglu. 2016.
\newblock \href {http://arxiv.org/abs/1610.10099} {Neural machine translation
  in linear time}.
\newblock \emph{CoRR}, abs/1610.10099.

\bibitem[{Kearnes et~al.(2016)Kearnes, McCloskey, Berndl, Pande, and
  Riley}]{kearnes2016molecular}
Steven Kearnes, Kevin McCloskey, Marc Berndl, Vijay Pande, and Patrick Riley.
  2016.
\newblock Molecular graph convolutions: moving beyond fingerprints.
\newblock \emph{Journal of computer-aided molecular design}, 30(8):595--608.

\bibitem[{Kingma and Ba(2015)}]{kingma2015adam}
Diederik~P. Kingma and Jimmy Ba. 2015.
\newblock \href {http://arxiv.org/abs/1412.6980} {Adam: {A} method for
  stochastic optimization}.
\newblock In \emph{ICLR}.

\bibitem[{Kipf and Welling(2016)}]{kipf2016semigraphconv}
Thomas~N. Kipf and Max Welling. 2016.
\newblock \href {http://arxiv.org/abs/1609.02907} {Semi-supervised
  classification with graph convolutional networks}.
\newblock \emph{CoRR}, abs/1609.02907.

\bibitem[{Linzen et~al.(2016)Linzen, Dupoux, and
  Goldberg}]{linzen-dupoux-goldberg:2016:tacllstmsyntax}
Tal Linzen, Emmanuel Dupoux, and Yoav Goldberg. 2016.
\newblock \href {https://www.transacl.org/ojs/index.php/tacl/article/view/972}
  {Assessing the ability of lstms to learn syntax-sensitive dependencies}.
\newblock \emph{Transactions of the Association for Computational Linguistics},
  4:521--535.

\bibitem[{Luong et~al.(2015{\natexlab{a}})Luong, Le, Sutskever, Vinyals, and
  Kaiser}]{luong16iclr}
Minh{-}Thang Luong, Quoc~V. Le, Ilya Sutskever, Oriol Vinyals, and Lukasz
  Kaiser. 2015{\natexlab{a}}.
\newblock \href {http://arxiv.org/abs/1511.06114} {{Multi-task Sequence to
  Sequence Learning}}.
\newblock \emph{CoRR}, abs/1511.06114.

\bibitem[{Luong et~al.(2015{\natexlab{b}})Luong, Pham, and
  Manning}]{luong15emnlp}
Thang Luong, Hieu Pham, and Christopher~D. Manning. 2015{\natexlab{b}}.
\newblock \href {http://aclweb.org/anthology/D15-1166} {{Effective Approaches
  to Attention-based Neural Machine Translation}}.
\newblock In \emph{{Proceedings of the 2015 Conference on Empirical Methods in
  Natural Language Processing}}, pages 1412--1421, Lisbon, Portugal.
  Association for Computational Linguistics.

\bibitem[{{Marcheggiani} and {Titov}(2017)}]{marcheggiani-titov:2017:srlgcn}
Diego {Marcheggiani} and Ivan {Titov}. 2017.
\newblock \href {http://arxiv.org/abs/1703.04826} {{Encoding Sentences with
  Graph Convolutional Networks for Semantic Role Labeling}}.
\newblock In \emph{{Proceedings of the 2017 Conference on Empirical Methods in
  Natural Language Processing}}, Copenhagen, Denmark. Association for
  Computational Linguistics.

\bibitem[{{Nadejde} et~al.(2017){Nadejde}, {Reddy}, {Sennrich}, {Dwojak},
  {Junczys-Dowmunt}, {Koehn}, and {Birch}}]{nadejdeetal2017syntaxaware}
Maria {Nadejde}, Siva {Reddy}, Rico {Sennrich}, Tomasz {Dwojak}, Marcin
  {Junczys-Dowmunt}, Philipp {Koehn}, and Alexandra {Birch}. 2017.
\newblock \href {http://arxiv.org/abs/1702.01147} {{Syntax-aware Neural Machine
  Translation Using CCG}}.
\newblock \emph{ArXiv e-prints}.

\bibitem[{Papineni et~al.(2002)Papineni, Roukos, Ward, and jing
  Zhu}]{papineni2002bleu}
Kishore Papineni, Salim Roukos, Todd Ward, and Wei jing Zhu. 2002.
\newblock Bleu: a method for automatic evaluation of machine translation.
\newblock pages 311--318.

\bibitem[{Schlichtkrull et~al.(2017)Schlichtkrull, Kipf, Bloem, Berg, Titov,
  and Welling}]{schlichtkrull2017}
Michael Schlichtkrull, Thomas~N Kipf, Peter Bloem, Rianne van~den Berg, Ivan
  Titov, and Max Welling. 2017.
\newblock \href {http://arxiv.org/abs/1703.06103} {{Modeling Relational Data
  with Graph Convolutional Networks}}.
\newblock \emph{ArXiv e-prints}.

\bibitem[{Schuster and Paliwal(1997)}]{schusterpaliwal1997}
Mike Schuster and Kuldip~K. Paliwal. 1997.
\newblock \href {https://doi.org/10.1109/78.650093} {{Bidirectional recurrent
  neural networks}}.
\newblock \emph{IEEE Transactions on Signal Processing}, 45(11):2673--2681.

\bibitem[{Sennrich and Haddow(2016)}]{sennrich2016linguistic}
Rico Sennrich and Barry Haddow. 2016.
\newblock \href {http://arxiv.org/abs/1606.02892} {{Linguistic Input Features
  Improve Neural Machine Translation}}.
\newblock In \emph{{Proceedings of the First Conference on Machine Translation
  (WMT16)}}, volume abs/1606.02892.

\bibitem[{Sennrich et~al.(2016{\natexlab{a}})Sennrich, Haddow, and
  Birch}]{sennrich-haddow-birch:2016:WMT}
Rico Sennrich, Barry Haddow, and Alexandra Birch. 2016{\natexlab{a}}.
\newblock \href {http://www.aclweb.org/anthology/W16-2323} {Edinburgh neural
  machine translation systems for wmt 16}.
\newblock In \emph{Proceedings of the First Conference on Machine Translation},
  pages 371--376, Berlin, Germany. Association for Computational Linguistics.

\bibitem[{Sennrich et~al.(2016{\natexlab{b}})Sennrich, Haddow, and
  Birch}]{sennrich2016subword}
Rico Sennrich, Barry Haddow, and Alexandra Birch. 2016{\natexlab{b}}.
\newblock \href {http://www.aclweb.org/anthology/P16-1162} {Neural machine
  translation of rare words with subword units}.
\newblock In \emph{Proceedings of the 54th Annual Meeting of the Association
  for Computational Linguistics (Volume 1: Long Papers)}, pages 1715--1725,
  Berlin, Germany. Association for Computational Linguistics.

\bibitem[{Shi et~al.(2016)Shi, Padhi, and Knight}]{shi2016emnlp}
Xing Shi, Inkit Padhi, and Kevin Knight. 2016.
\newblock \href {https://aclweb.org/anthology/D16-1159} {Does string-based
  neural mt learn source syntax?}
\newblock In \emph{Proceedings of the 2016 Conference on Empirical Methods in
  Natural Language Processing}, pages 1526--1534, Austin, Texas. Association
  for Computational Linguistics.

\bibitem[{Smith and Eisner(2006)}]{smith-eisner:2006:WMT}
David Smith and Jason Eisner. 2006.
\newblock \href {http://www.aclweb.org/anthology/W/W06/W06-3104}
  {Quasi-synchronous grammars: Alignment by soft projection of syntactic
  dependencies}.
\newblock In \emph{Proceedings on the Workshop on Statistical Machine
  Translation}, pages 23--30, New York City. Association for Computational
  Linguistics.

\bibitem[{Snover et~al.(2006)Snover, Dorr, Schwartz, Micciulla, and
  Makhoul}]{snover2006ter}
Matthew Snover, Bonnie Dorr, Richard Schwartz, Linnea Micciulla, and John
  Makhoul. 2006.
\newblock A study of translation edit rate with targeted human annotation.
\newblock In \emph{In Proceedings of Association for Machine Translation in the
  Americas}, pages 223--231.

\bibitem[{Socher et~al.(2013)Socher, Perelygin, Wu, Chuang, Manning, Ng, and
  Potts}]{socher-EtAl:2013:EMNLP}
Richard Socher, Alex Perelygin, Jean Wu, Jason Chuang, Christopher~D. Manning,
  Andrew Ng, and Christopher Potts. 2013.
\newblock \href {http://www.aclweb.org/anthology/D13-1170} {Recursive deep
  models for semantic compositionality over a sentiment treebank}.
\newblock In \emph{Proceedings of EMNLP}.

\bibitem[{Stahlberg et~al.(2016)Stahlberg, Hasler, Waite, and
  Byrne}]{stahlberg2016syntactically}
Felix Stahlberg, Eva Hasler, Aurelien Waite, and Bill Byrne. 2016.
\newblock \href {http://anthology.aclweb.org/P16-2049} {Syntactically guided
  neural machine translation}.
\newblock In \emph{Proceedings of the 54th Annual Meeting of the Association
  for Computational Linguistics (Volume 2: Short Papers)}, pages 299--305,
  Berlin, Germany. Association for Computational Linguistics.

\bibitem[{Stanojevi{\'c} and Sima’an(2015)}]{stanojevic2015evaluating}
Milo{\v{s}} Stanojevi{\'c} and Khalil Sima’an. 2015.
\newblock Evaluating mt systems with beer.
\newblock \emph{The Prague Bulletin of Mathematical Linguistics},
  104(1):17--26.

\bibitem[{Stanojevi\'{c} and Sima'an(2014)}]{stanojevic-simaan:2014:EMNLP2014}
Milo\v{s} Stanojevi\'{c} and Khalil Sima'an. 2014.
\newblock \href {http://www.aclweb.org/anthology/D14-1025} {Fitting sentence
  level translation evaluation with many dense features}.
\newblock In \emph{Proceedings of the 2014 Conference on Empirical Methods in
  Natural Language Processing (EMNLP)}, pages 202--206, Doha, Qatar.
  Association for Computational Linguistics.

\bibitem[{Surdeanu et~al.(2008)Surdeanu, Johansson, Meyers, M{\`{a}}rquez, and
  Nivre}]{surdeanuJMMN08}
Mihai Surdeanu, Richard Johansson, Adam Meyers, Llu{\'{\i}}s M{\`{a}}rquez, and
  Joakim Nivre. 2008.
\newblock \href {http://aclweb.org/anthology/W/W08/W08-2121.pdf} {The conll
  2008 shared task on joint parsing of syntactic and semantic dependencies}.
\newblock In \emph{Proceedings of CoNLL}.

\bibitem[{Sutskever et~al.(2014)Sutskever, Vinyals, and
  Le}]{sutskever2014seq2seqnips}
Ilya Sutskever, Oriol Vinyals, and Quoc~V. Le. 2014.
\newblock \href
  {http://papers.nips.cc/paper/5346-sequence-to-sequence-learning-with-neural-networks}
  {{Sequence to Sequence Learning with Neural Networks}}.
\newblock In \emph{{Neural Information Processing Systems (NIPS)}}, pages
  3104--3112.

\bibitem[{Wu et~al.(2016)Wu, Schuster, Chen, Le, Norouzi, Macherey, Krikun,
  Cao, Gao, Macherey, Klingner, Shah, Johnson, Liu, Kaiser, Gouws, Kato, Kudo,
  Kazawa, Stevens, Kurian, Patil, Wang, Young, Smith, Riesa, Rudnick, Vinyals,
  Corrado, Hughes, and Dean}]{google2016nmt}
Yonghui Wu, Mike Schuster, Zhifeng Chen, Quoc~V. Le, Mohammad Norouzi, Wolfgang
  Macherey, Maxim Krikun, Yuan Cao, Qin Gao, Klaus Macherey, Jeff Klingner,
  Apurva Shah, Melvin Johnson, Xiaobing Liu, Lukasz Kaiser, Stephan Gouws,
  Yoshikiyo Kato, Taku Kudo, Hideto Kazawa, Keith Stevens, George Kurian,
  Nishant Patil, Wei Wang, Cliff Young, Jason Smith, Jason Riesa, Alex Rudnick,
  Oriol Vinyals, Greg Corrado, Macduff Hughes, and Jeffrey Dean. 2016.
\newblock \href {http://arxiv.org/abs/1609.08144} {Google's neural machine
  translation system: Bridging the gap between human and machine translation}.
\newblock \emph{CoRR}, abs/1609.08144.

\bibitem[{Yogatama et~al.(2016)Yogatama, Blunsom, Dyer, Grefenstette, and
  Ling}]{YogatamaBDGL16}
Dani Yogatama, Phil Blunsom, Chris Dyer, Edward Grefenstette, and Wang Ling.
  2016.
\newblock \href {http://arxiv.org/abs/1611.09100} {Learning to compose words
  into sentences with reinforcement learning}.
\newblock \emph{CoRR}, abs/1611.09100.

\bibitem[{Zollmann and Venugopal(2006)}]{Zollmann:2006:SAMT}
Andreas Zollmann and Ashish Venugopal. 2006.
\newblock \href {http://dl.acm.org/citation.cfm?id=1654650.1654671} {Syntax
  augmented machine translation via chart parsing}.
\newblock In \emph{Proceedings of the Workshop on Statistical Machine
  Translation}, StatMT '06, pages 138--141, Stroudsburg, PA, USA. Association
  for Computational Linguistics.

\end{thebibliography}
